\documentclass[11pt]{article}

\usepackage{acl}
\usepackage{times}
\usepackage{latexsym}

\usepackage[T1]{fontenc}

\usepackage[utf8]{inputenc}

\usepackage{microtype}

\usepackage{inconsolata}

\usepackage{graphicx}
\usepackage{hyperref}
\usepackage{url}
\usepackage{algorithm}
\usepackage{algorithmic}
\usepackage{amsmath}
\usepackage{amssymb}
\usepackage{enumitem}
\usepackage{booktabs}
\usepackage{subcaption}
\usepackage{tabularx}
\usepackage{nicematrix}
\usepackage{array}
\usepackage{geometry}
\usepackage{multirow}
\usepackage[table]{xcolor} 

\title{HeteroCache: A Dynamic Retrieval Approach to Heterogeneous KV Cache Compression for Long-Context LLM Inference}

\author{
  \textbf{Zhiyuan Shi\textsuperscript{1}}, 
  \textbf{Qibo Qiu\textsuperscript{2,3}}, 
  \textbf{Feng Xue\textsuperscript{4}}, 
  \textbf{Zhonglin Jiang\textsuperscript{4}}, \\
  \textbf{Li Yu\textsuperscript{2}}, 
  \textbf{Jian Jiang\textsuperscript{2}}, 
  \textbf{Xiaofei He\textsuperscript{3,5}}, 
  \textbf{Wenxiao Wang\textsuperscript{1,\textdagger}} \\
  \textsuperscript{1}School of Software Technology, Zhejiang University \\
  \textsuperscript{2}China Mobile (Zhejiang) Research \& Innovation Institute \\
  \textsuperscript{3}State Key Lab of CAD\&CG, Zhejiang University \\
  \textsuperscript{4}Geely Automobile Research Institute (Ningbo) Co., Ltd.  \hspace{1em}
  \textsuperscript{5}FABU Inc. \\
  \small{
   \texttt{\{zhiyuan0206, wenxiaowang\}@zju.edu.cn}
  }
}

\begin{document}
\maketitle

\begingroup
\def\thefootnote{\textdagger}
\footnotetext{Corresponding author}
\endgroup

\begin{abstract}
The linear memory growth of the KV cache poses a significant bottleneck for LLM inference in long-context tasks. Existing static compression methods often fail to preserve globally important information. Although recent dynamic retrieval approaches attempt to address this issue, they typically suffer from coarse-grained caching strategies and incur high I/O overhead. To overcome these limitations, we propose \textbf{HeteroCache}, a training-free dynamic compression framework. Our method is built on two key insights: attention heads exhibit diverse temporal heterogeneity, and there is significant spatial redundancy among heads within the same layer.
Guided by these insights, HeteroCache categorizes heads based on stability and similarity, applying a fine-grained weighting strategy that allocates larger cache budgets to heads with rapidly shifting attention to capture context changes.
Furthermore, it features a hierarchical storage mechanism where representative heads monitor attention drift to trigger asynchronous, on-demand context retrieval, thereby hiding I/O latency.
Experiments demonstrate that HeteroCache achieves state-of-the-art performance on long-context benchmarks and accelerates decoding by up to $3\times$ compared to the original model with a 224K context. Our code is available at \url{https://github.com/ponytaill/HeteroCache}.
\end{abstract}

\section{Introduction}
LLMs have reshaped artificial intelligence, demonstrating powerful capabilities in conversational assistants and autonomous agents \citep{singh2025openaigpt5card,yang2025qwen3technicalreport,geminiteam2025geminifamilyhighlycapable,wang2025openhandsopenplatformai}. This success depends on the processing of long-form context, urging innovative architectures to extend current limits \citep{deepseekai2025deepseekv3technicalreport,ye2025flashinferefficientcustomizableattention}. However, Transformer-based models \citep{vaswani2017attention} rely on an optimization mechanism known as the KV cache during generation. 
Although this avoids redundant computation, its linear memory consumption makes the KV cache bandwidth a major bottleneck for efficient inference on resource-constrained hardware.

To address this problem, some work highlights the inherent sparsity of the attention mechanism in long contexts, such as MInference \citep{10.5555/3737916.3739579}, which shows that only 4K tokens of 128K can account for 96.4\% of the total attention weight.
Consequently, a major line of research has focused on static compression, which permanently evicts unimportant token entries from the cache. Methods like SnapKV and PyramidKV \citep{li2024snapkv,cai2025pyramidkvdynamickvcache} typically utilize heuristics based on historical attention scores to identify and retain only a critical subset of tokens. However, this irreversible discard policy poses a fundamental risk. Due to attention drift in complex reasoning tasks, information deemed unimportant early on may become pivotal later, leading to accuracy degradation. To mitigate this, recent work such as ShadowKV and OmniKV \citep{sun2025shadowkvkvcacheshadows,hao2025omnikv} has introduced dynamic compression, effectively improving performance by recalling the necessary tokens. Nevertheless, these approaches face two primary limitations: first, they often employ coarse-grained caching strategies that overlook the heterogeneity across layers or heads; second, their coarse-grained retrieval at every step incurs unnecessary I/O overhead and potential accuracy drops.

To overcome these limitations, we propose HeteroCache, a dynamic framework that avoids premature information loss. Our approach leverages two key observations: temporal heterogeneity, where some heads maintain long-term stability while others exhibit rapid shifts, and intralayer redundancy, where attention patterns of certain heads are highly similar within the same layer. Guided by these insights, HeteroCache optimizes the compression through a fine-grained, role-based strategy.

First, we employ a profiling strategy that categorizes heads into distinct functional roles. By explicitly modeling their similarity behavior, we cluster highly redundant heads to identify representative ones, subsequently dividing all heads into the full heads that retain complete context and the compressed heads designated for compression. For compressed heads, we implement a fine-grained stability-based strategy that allocates larger cache budgets to heads prone to rapid context shifts. This ensures the precise capture of dynamic information and maintains stable model performance across varying contexts. Second, we ensure efficient inference through a hierarchical storage mechanism with sparse monitoring. Rather than retrieving context at every step, we offload the majority of the KV cache to CPU memory, retaining full context on the GPU only for the full heads. These heads continuously monitor attention drift and trigger asynchronous on-demand retrieval only when a significant attention shift is detected, utilizing data fetched from the CPU to dynamically update the compressed heads. This approach maintains high fidelity with minimal overhead.

Our extensive experiments were conducted on mainstream LLMs, including the standard Llama and Qwen series, as well as the reasoning-specialized DeepSeek-R1-Distill-Llama-8B to assess performance on chain-of-thought tasks. We evaluated these models under various compression rates on multiple long-context benchmarks, such as LongBench \citep{bai-etal-2024-longbench}, LongBench v2 \citep{bai-etal-2025-longbench}, InfiniteBench \citep{zhang-etal-2024-bench}. The results demonstrate that, compared to other advanced compression algorithms, HeteroCache achieves state-of-the-art performance on a variety of tasks while realizing up to 3× inference acceleration compared to the baseline of the original model with a 224K context.
\begin{figure*}[t]
    \centering
    \begin{subfigure}[b]{0.32\textwidth}
        \centering
        \includegraphics[width=\linewidth, height=4cm]{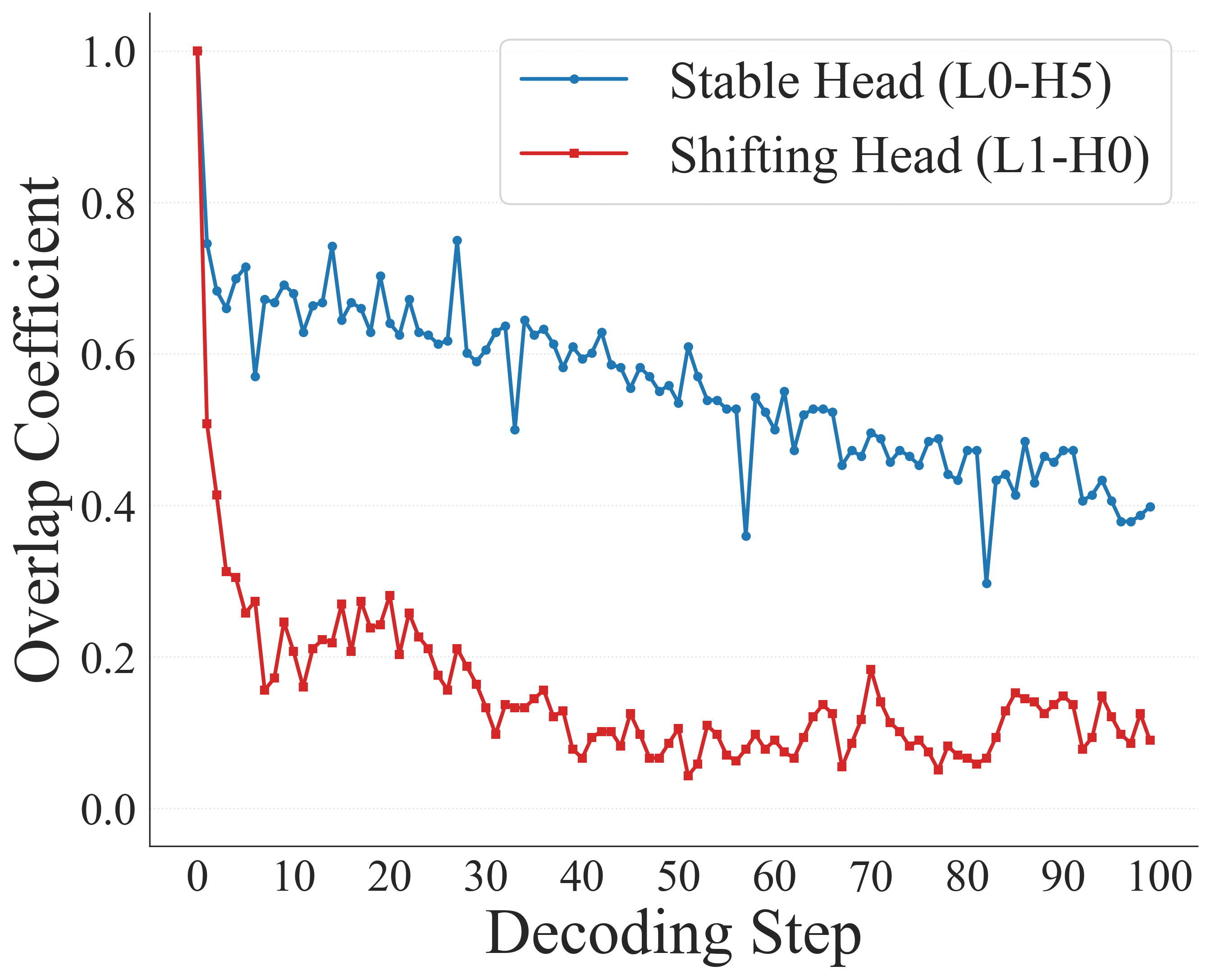} 
        \caption{Temporal Heterogeneity}
        \label{fig:method_a}
    \end{subfigure}
    \hfill
    \begin{subfigure}[b]{0.32\textwidth}
        \centering
        \includegraphics[width=\linewidth, height=4cm]{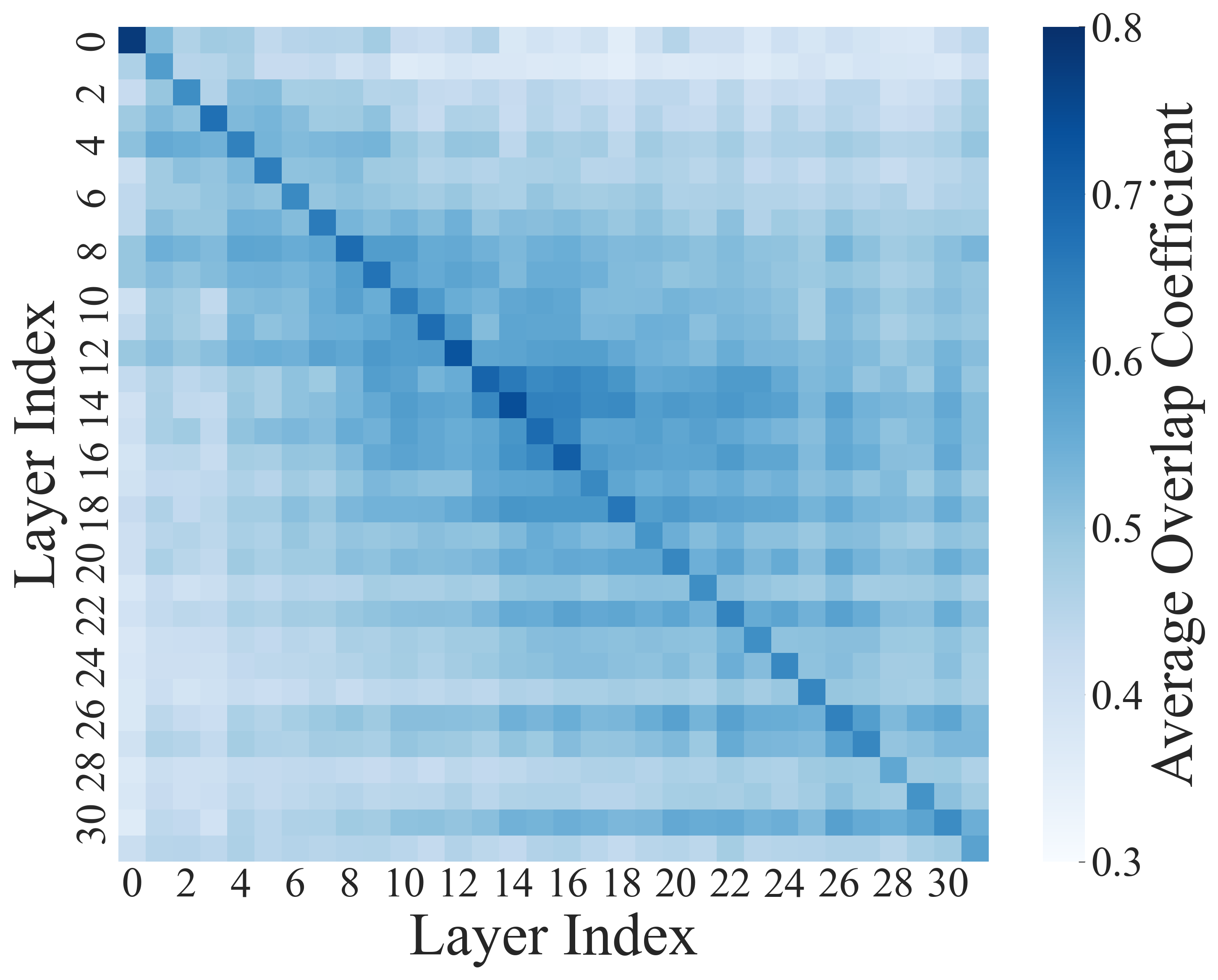}
        \caption{Intralayer Redundancy}
        \label{fig:method_b}
    \end{subfigure}
    \hfill
    \begin{minipage}[b]{0.25\textwidth}
        \begin{subfigure}[b]{\linewidth}
            \centering
            \includegraphics[width=\linewidth, height=2cm]{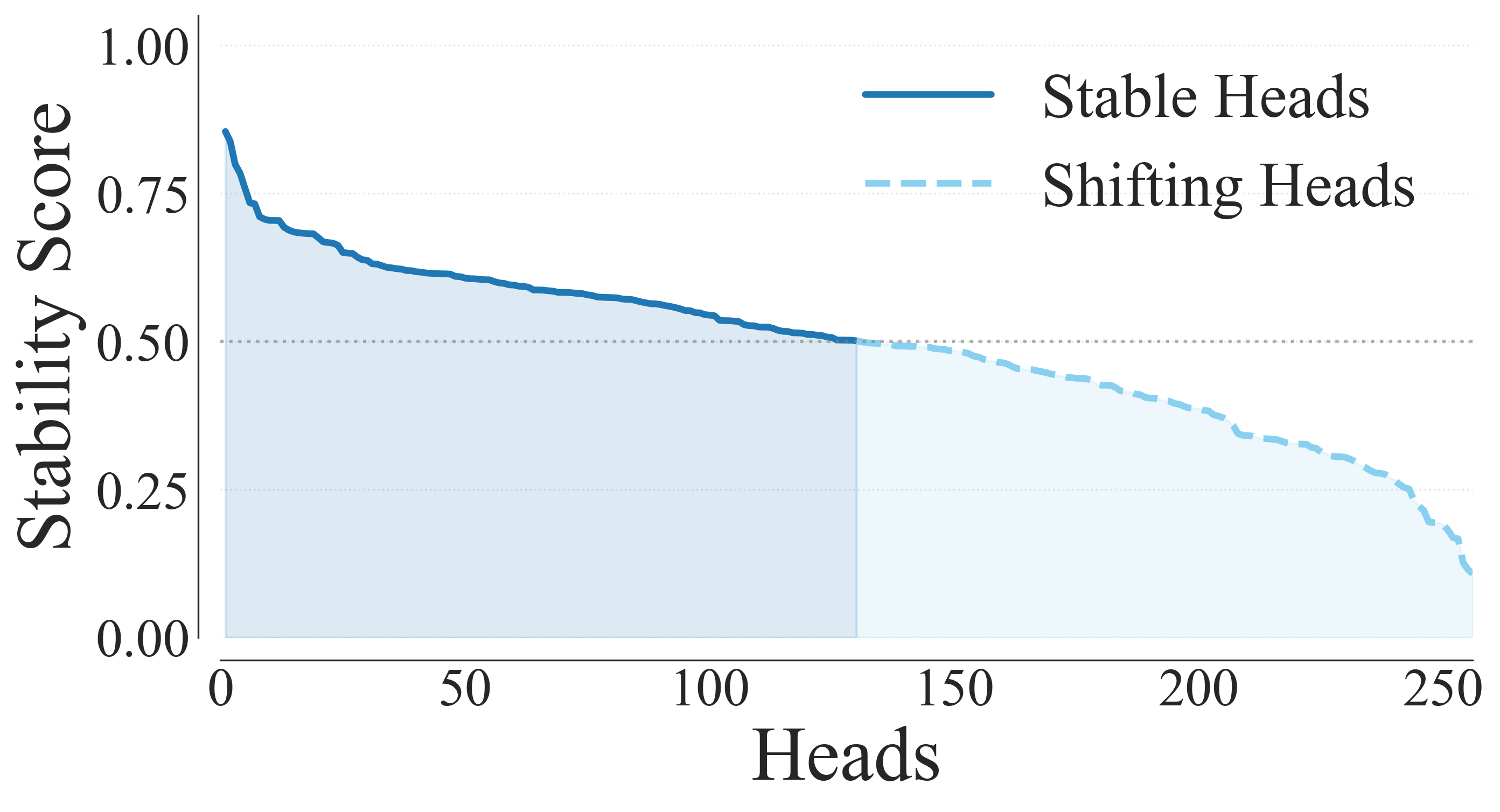}
            \caption{Stability Classification}
            \label{fig:detail_1}
        \end{subfigure}
        \begin{subfigure}[b]{\linewidth}
            \centering
            \includegraphics[width=\linewidth, height=2cm]{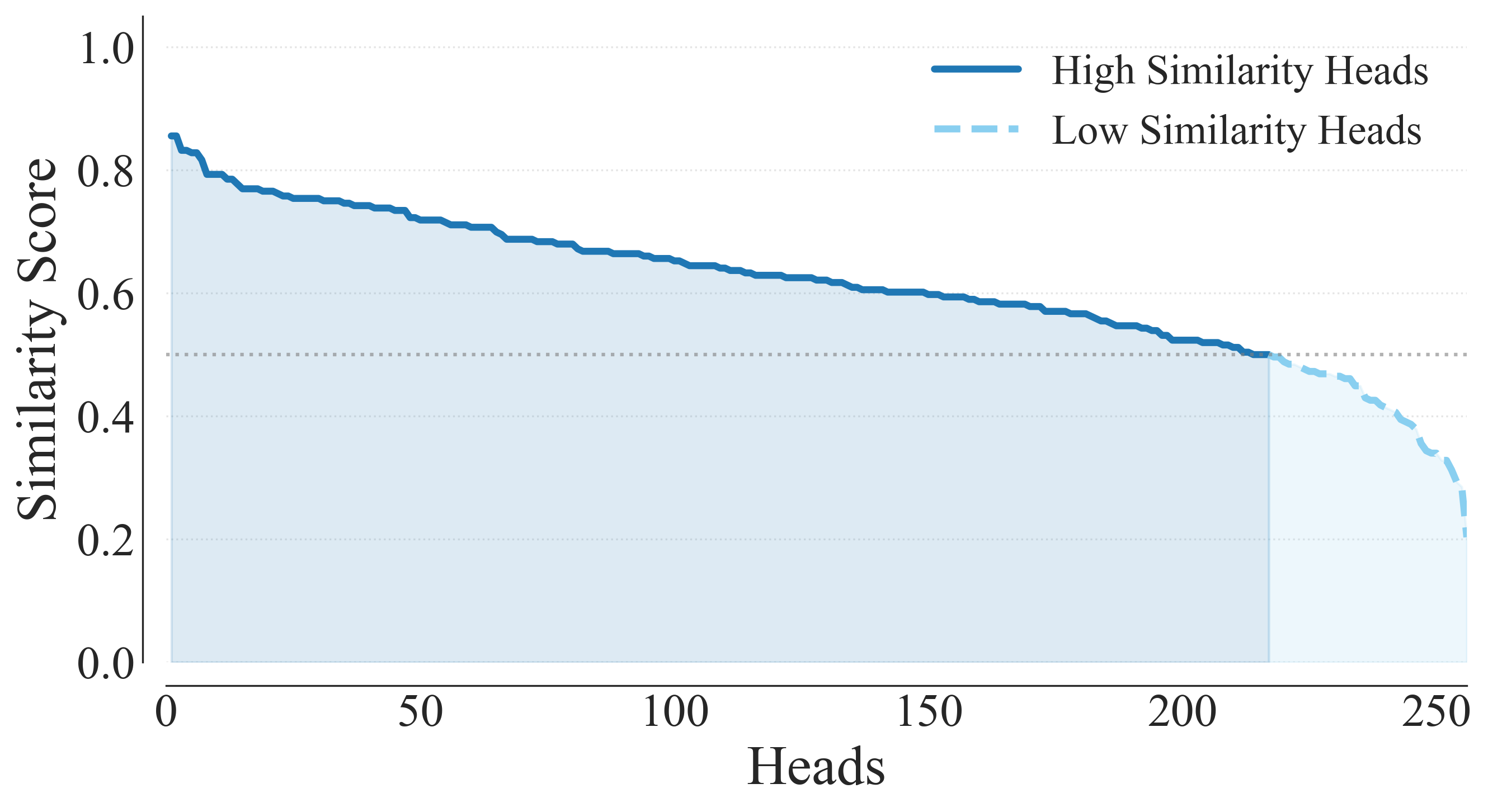}
            \caption{Similarity Classification}
            \label{fig:detail_2}
        \end{subfigure}
    \end{minipage}
    \caption{Analysis of attention heads heterogeneity and redundancy. (a) illustrates temporal heterogeneity: the \textcolor{blue}{blue line} represents a stable head maintaining consistent focus, while the \textcolor{red}{red line} indicates a shifting head with rapid attention shift; (b) visualizes intralayer redundancy, where the darker diagonal blocks indicate that attention heads within the same layer share significantly higher similarity than those across different layers; (c) and (d) present the classification of heads based on their stability and similarity scores, respectively.}
    \label{fig:observation}
\end{figure*}

\section{Related Work}
\paragraph{KV Cache Eviction.}
Existing work has demonstrated the inherent sparsity of the attention mechanism in long contexts \citep{10.5555/3737916.3739579}. Therefore, many KV cache eviction works are studying how to effectively retain important tokens in the inference stage and evict unimportant ones. StreamingLLM \citep{xiao2023streamingllm} relied on a simple positional heuristic to retain initial and final tokens. This simple eviction strategy leads to significant information loss. To address this, subsequent methods introduced more sophisticated indicators driven by attention to identify less important tokens. H2O \citep{10.5555/3666122.3667628} uses cumulative attention to eliminate the last token. SnapKV \citep{li2024snapkv} uses window attention clustering in the prefill stage to obtain the set of important tokens. CAKE \citep{qin2025cake} introduces a cascading and adaptive strategy that evaluates layer-specific preferences to rationally distribute cache resources. 

Nevertheless, the static methods above suffer from a critical limitation. Tokens that are regarded as irrelevant and evicted at a certain stage cannot be retrieved, which may become important in subsequent processes, resulting in significant information loss. 
Some works have recognized the importance of dynamic selection, such as Quest \citep{tang2024quest}. 
However, it fails to reduce memory usage and suffers from accuracy degradation.
ShadowKV \citep{sun2025shadowkvkvcacheshadows} and OmniKV \citep{hao2025omnikv} attempt to handle infinite context by offloading KV pairs to CPU memory. Unfortunately, their coarse-grained retrieval strategies often overlook fine-grained dependencies and incur high I/O overhead.

Some methods, such as KVQuant \citep{hooper2024kvquant} and KIVI \citep{liu2024kivi}, attempt to quantize the KV cache.
These quantization methods are orthogonal to the token-level eviction strategy of the KV cache and can be combined to further substantially reduce GPU memory overhead.

\begin{figure*}[t]
    \centering

    \includegraphics[width=\textwidth]{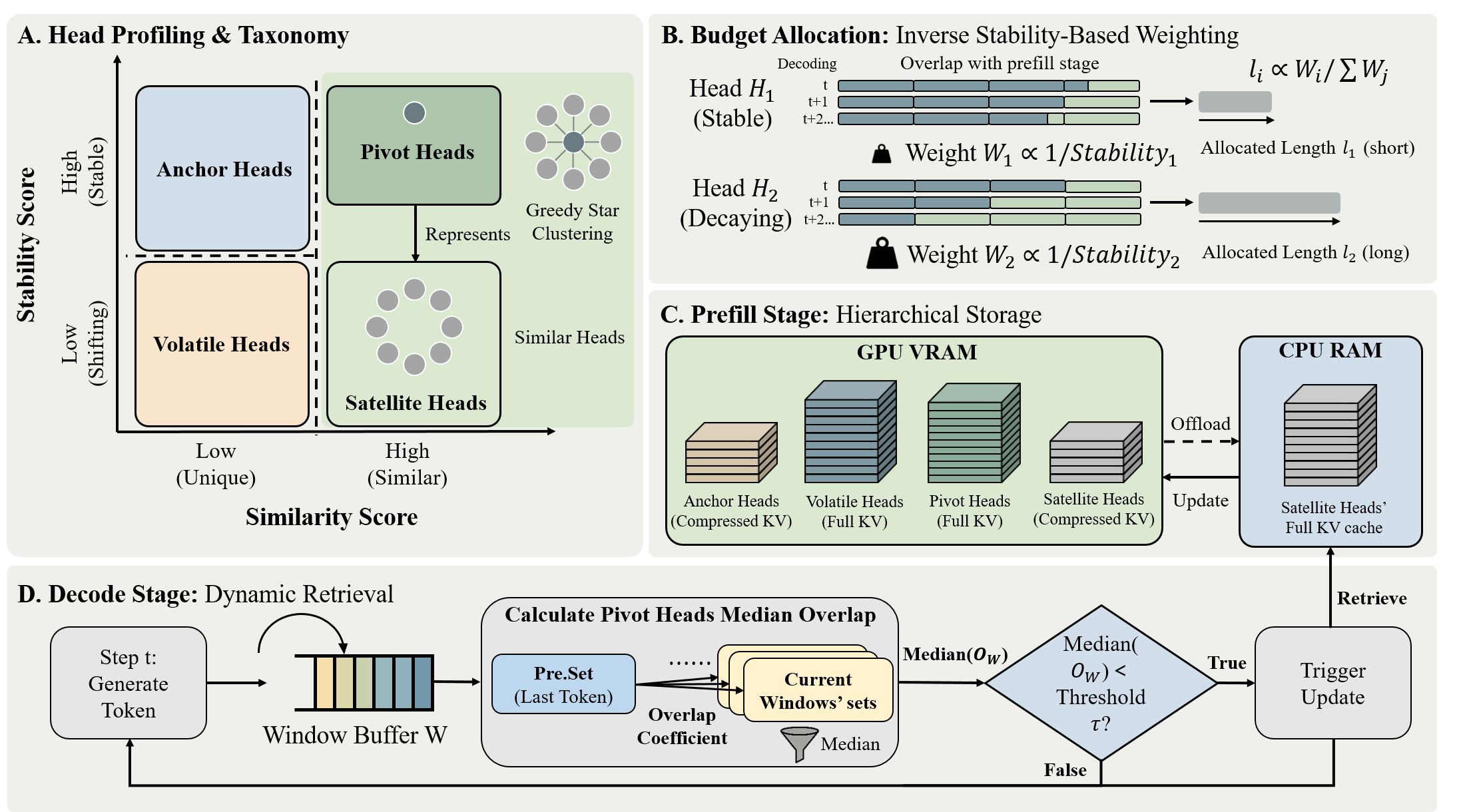}

    \caption{The workflow of HeteroCache. \textbf{(A, B) Offline Calibration:} Heads are categorized into functional roles (A) to determine stability-based budgets for compressed heads (B). \textbf{(C, D) Online Inference:} Guided by (B), (C) initializes hierarchical storage in prefill stage. In (D), Pivot heads monitor drift to trigger asynchronous CPU retrieval for updating satellite heads in decode stage.}
    \label{fig:heterocache}
\end{figure*}

\section{Observation}
\label{sec:observation}
To investigate the intrinsic attention behavior of the KV cache in long-context inference, we conducted a pilot study using a small calibration dataset. More details on the data collection process and visualization methodology are provided in the Appendix~\ref{sec:appendix_b}. We focus on quantifying the behavior of attention heads across two dimensions: temporal heterogeneity and intralayer redundancy. Specifically, we employ the overlap coefficient \citep{szymkiewicz1934conlribution} to evaluate the alignment between the sets of the top-$k$ important indices from any two sources, $\mathcal{K}(X)$ and $\mathcal{K}(Y)$, calculated as follows:
\begin{equation}
\small
\text{O}(\mathcal{K}(X), \mathcal{K}(Y)) = \frac{|\mathcal{K}(X) \cap \mathcal{K}(Y)|}{\min(|\mathcal{K}(X)|, |\mathcal{K}(Y)|)}
\end{equation}
where $X$ and $Y$ represent attention scores (e.g., from different decoding steps or different attention heads) and $\mathcal{K}(\cdot)$ denotes the set of indices corresponding to the top-$k$ values. This metric serves as a proxy for information, where a high overlap indicates that the head persistently attends to the same context.

\paragraph{Temporal Heterogeneity.}
We first examine how the focus of a single attention head evolves over time. By tracking the overlap coefficient between the top-$k$ attended tokens in the decode stage and the prefill stage over a span of $T$ steps (e.g., $T=100$), we observe distinct behaviors. As shown in Figure~\ref{fig:observation} (a), certain heads exhibit slow decay, where the set of important tokens remains highly consistent with the prefill phase, indicating a persistent focus on specific contexts. Conversely, some heads show fast decay with a rapid decline in overlap, suggesting that their attention focus shifts dynamically as new tokens are generated. To quantify this property, we define the stability score $S_{\text{stable}}^{(h)}$ for a head $h$ as the median of its overlap coefficients with the prefill stage across $T$ decoding steps:

\begin{equation}
\small
S_{\text{stable}}^{(h)} = \operatorname*{Median}_{t=1 \dots T} \left( \text{O}(\mathcal{K}_{t}^{(h)}, \mathcal{K}_{\text{prefill}}^{(h)}) \right)
\end{equation}

Figure~\ref{fig:observation}~(c) shows the stability scores of all heads ranked in descending order. Based on a predefined stability threshold $\tau_{\text{stable}}$, we classify the heads into stable heads where $S_{\text{stable}} \ge \tau_{\text{stable}}$ and shifting heads where $S_{\text{stable}} < \tau_{\text{stable}}$.

\paragraph{Intralayer Redundancy.}
In addition to the temporal heterogeneity of attention, we also investigated spatial redundancy across attention heads.  To quantify spatial relationships, we define layer-level similarity as the average of the overlap coefficients from each head in a source layer to a target layer. The result, shown in Figure~\ref{fig:observation} (b), reveals a strong intralayer similarity, which is significantly higher than the interlayer similarity.
This high degree of intralayer redundancy suggests that not all heads contribute unique information; many are functionally similar.
To measure this redundancy, we calculate the similarity score $S_{\text{sim}}^{(h)}$. For each step $t$, we first identify the maximum overlap between the head $h$ and all other heads $h'$ in the same layer $l$. The final score is the median of these maximums in the $T$ steps:

\begin{equation}
\small
S_{\text{sim}}^{(h)} = \operatorname*{Median}_{t=1 \dots T} \left( \max_{h' \in l} \text{O}(\mathcal{K}_{t}^{(h)}, \mathcal{K}_{t}^{(h')}) \right)
\end{equation}

We plot the heads sorted by their similarity scores in Figure~\ref{fig:observation}~(d). Specifically, based on similarity threshold $\tau_{\text{sim}}$, we designate heads as similar heads where $S_{\text{sim}} \ge \tau_{\text{sim}}$, implying that their information can be approximated by others, and conversely as unique heads where $S_{\text{sim}} < \tau_{\text{sim}}$. Collectively, these two dimensions form the basis for our heterogeneous taxonomy strategy in HeteroCache.

\section{Method}
\label{sec:method}

Building on the observations of temporal heterogeneity and intralayer redundancy presented in Section~\ref{sec:observation}, we propose HeteroCache, a framework designed to compress the KV cache through heterogeneous head profiling and dynamic retrieval. The detailed pseudocode for our algorithm is provided in Appendix~\ref{sec:pseudocode}. This section details the three phases of our approach: head profiling and taxonomy, stability-based budget allocation, and hierarchical storage with dynamic retrieval.

\subsection{Head Profiling and Taxonomy}
\label{sec:taxonomy}

To systematically exploit the heterogeneity of attention heads, we introduce an offline calibration phase to compute the stability score $S_{\text{stable}}^{(h)}$ and similarity score $S_{\text{sim}}^{(h)}$ for each head $h$. Based on these metrics, we categorize the set of all attention heads $\mathcal{H}$ into four functional roles, as shown in Figure~\ref{fig:heterocache} (A). We first differentiate heads according to the similarity threshold $\tau_{\text{sim}}$, defining the sets of unique heads and similar heads as:
\begin{equation}
\small
\begin{split}
    \mathcal{H}_{\text{unique}} &= \{h \in \mathcal{H} \mid S_{\text{sim}}^{(h)} < \tau_{\text{sim}}\} \\
    \mathcal{H}_{\text{similar}} &= \{h \in \mathcal{H} \mid S_{\text{sim}}^{(h)} \ge \tau_{\text{sim}}\}
\end{split}
\end{equation}

The unique heads are characterized by their specialized functional roles, exhibiting attention patterns that diverge significantly from other heads. Given their distinct nature, we further classify them based on their stability scores. Using the stability threshold $\tau_{\text{stable}}$, we delineate anchor heads and volatile heads as:
\begin{equation}
\small
\begin{split}
    \mathcal{H}_{\text{anchor}} = \{h \in \mathcal{H}_{\text{unique}} \mid S_{\text{stable}}^{(h)} \ge \tau_{\text{stable}}\} \\
    \mathcal{H}_{\text{volatile}} = \{h \in \mathcal{H}_{\text{unique}} \mid S_{\text{stable}}^{(h)} < \tau_{\text{stable}}\}
\end{split}
\end{equation}
Anchor heads maintain a consistent focus on specific contexts, making their history highly predictable and suitable for compression. In contrast, volatile heads exhibit rapid attention shift; therefore, we retain their full KV cache on the GPU to prevent information loss. 
Empirically, volatile heads are sparse and their proportion is tunable through $\tau_{\text{stable}}$, ensuring that they do not exhaust the memory budget.

For similar heads $\mathcal{H}_{\text{similar}}$, characterized by high redundancy, we employ the greedy star clustering algorithm \citep{Aslam_Pelekhov_Rus_2004} to divide them into clusters $\mathcal{C} = \{C_1, \dots, C_m\}$, ensuring that every head in a cluster is similar to its central head. Within each cluster $C_j$, the head with the highest centrality is designated as the pivot head $h_{\text{pivot}}^{(j)}$. Since the attention distribution of the pivot head effectively approximates that of the other members in the cluster, we retain its full KV cache to serve as a representative monitor for the attention shift. The remaining heads are classified as satellite heads:
\begin{equation}
\small
    \mathcal{H}_{\text{satellite}} = \bigcup_j (C_j \setminus \{h_{\text{pivot}}^{(j)}\})
\end{equation} where $\setminus$ denotes the set difference operation.
Consequently, satellite heads undergo compression to significantly minimize GPU memory usage, yet remain subject to dynamic updates guided by the pivot heads during inference. For a comprehensive description of the implementation pipeline, in conjunction with evaluations of the robustness, efficiency, and transferability of our offline calibration, see the Appendix~\ref{sec:appendix_c}.

\subsection{Stability-Based Budget Allocation}
\label{sec:budget}

Although numerous KV cache compression algorithms \citep{cai2025pyramidkvdynamickvcache,qin2025cake} have demonstrated varying degrees of redundancy across model layers, few approaches utilize fine-grained compression at the individual attention head level. Using the functional head classification of the previous section, we achieve this granularity by incorporating both the anchor and satellite heads into the compressed head set defined as $\mathcal{H}_{\text{comp}} = \mathcal{H}_{\text{anchor}} \cup \mathcal{H}_{\text{satellite}}$. Our intuition is that heads with higher stability scores exhibit concentrated attention regions, which permit a reduced budget, whereas heads with lower stability display shifting attention patterns that necessitate a larger allocation. To address this, we propose a fine-grained inverse stability-based weighting strategy, as illustrated in Figure~\ref{fig:heterocache} (B), where we assign a weight $w_i = 1 / S_{\text{stable}}^{(i)}$ to each compressed head $h_i \in \mathcal{H}_{\text{comp}}$ inversely proportional to its stability score. 
To determine the allocated KV cache length $l_i$ for each compressed head, we first calculate a compressed base length $L_{\text{base}}$. Let $\rho$ be the KV cache budget ratio, $N$ be the total number of heads, and $L$ be the sequence length. The base length $L_{\text{base}}$ is calculated as:
\begin{equation}
\small
L_{\text{base}} = \frac{(\rho N - N_{\text{full}})L}{N_{\text{comp}}}
\end{equation}
Given the total budget $N_{\text{comp}} \cdot L_{\text{base}}$, $l_i$ is:
\begin{equation}
\small
    l_i = (N_{\text{comp}} \cdot L_{\text{base}}) \cdot \frac{w_i}{\sum_{h_j \in \mathcal{H}_{\text{comp}}} w_j}
\end{equation}
This inverse weighting allows HeteroCache to adaptively safeguard context-sensitive heads against premature eviction.

\subsection{Hierarchical Storage \& Retrieval}
\label{sec:thirdpart}
To efficiently manage memory consumption, HeteroCache implements a hierarchical storage mechanism bridging GPU VRAM and CPU RAM, as illustrated in Figure~\ref{fig:heterocache} (C).

During the prefill stage, we maintain the full KV caches for the $\mathcal{H}_{\text{full}}$ volatile and pivot heads on the GPU. For compressed heads $\mathcal{H}_{\text{comp}}$, which comprise both anchor heads and satellite heads, we adopt a split storage strategy. 
Based on Section~\ref{sec:budget}, we assign a specific budget $l_i$ to each compressed head. In the GPU VRAM, we only retain the KV cache corresponding to the top-$l_i$ attention scores for these heads. Simultaneously, the full KV contexts of the satellite heads are offloaded to the CPU RAM to serve as a retrieval reservoir.

We formally describe the dynamic retrieval process for satellite heads as follows, also shown in Figure~\ref{fig:heterocache} (D). Let $h_p$ be a pivot head. We initialize a baseline set $\mathcal{K}_{\text{base}}$ consisting of the top-$L_{\text{base}}$ attended indices of the prefill stage. During the decoding step $t$, the pivot head monitors the shift of the attention distribution by calculating the overlap coefficient $o_t$ utilizing the base length $L_{\text{base}}$ as normalization factor:
\begin{equation}
\small
o_t = \text{O}(\mathcal{K}_t, \mathcal{K}_{\text{base}}) = \frac{|\mathcal{K}_t \cap \mathcal{K}_{\text{base}}|}{L_{\text{base}}}
\end{equation}
To differentiate meaningful attention shift from transient noise, we employ a sliding window of size $W$. A retrieval signal $r_t \in \{0, 1\}$ is triggered when the median overlap within the window falls below a predefined drift threshold $\tau_{\text{drift}}$:
\begin{equation}
\small
r_t = \mathbb{I}\left( \operatorname*{Median}\left(\{o_i\}_{i=t-W+1}^{t}\right) < \tau_{\text{drift}} \right)
\end{equation}
Upon detecting a drift where $r_t = 1$, the system triggers an update guided by the pivot head.

Specifically, we utilize the indices of the top-$l_{i}$ tokens from the pivot head's current attention distribution to retrieve the corresponding KV tensors from the CPU, updating the satellite heads on the GPU. Subsequently, we update the historical baseline state to prepare for re-initiating the next dynamic retrieval. Since this process is executed asynchronously, I/O latency is effectively hidden within the attention computation overhead, ensuring efficient inference without stalling. 
Additionally, anchor heads maintain their statically compressed state without compromising performance, benefiting from their high temporal stability.

\begin{table*}[t]
\centering 
\setlength{\tabcolsep}{4pt} 
\definecolor{baselinegray}{gray}{0.9}
\resizebox{\textwidth}{!}{%
\begin{NiceTabular}{l | c | c | c c c c c c}[colortbl-like]
\toprule 
\textbf{Methods} & \textbf{Mem\%} & \textbf{Avg.} & \textbf{Single-Doc QA} & \textbf{Multi-Doc QA} & \textbf{Summarize} & \textbf{Few-Shot} & \textbf{Synthetic} & \textbf{Code} \\
\midrule 
\rowcolor{baselinegray} \textit{Llama-3.1-8B-Instruct} & 100\% & 49.77 & 43.53 & 40.82 & 29.04 & 69.48 & 53.75 & 62.01 \\
Quest & 100\% & 48.60 & 42.79 & 40.12 & \textbf{29.11} & 67.50 & 52.52 & 59.55 \\
ShadowKV & 50\% & 48.03 & 42.10 & 40.42 & 23.34 & \textbf{69.48} & \textbf{53.75} & 59.08 \\
OmniKV & 50\% & 49.22 & 43.28 & 40.33 & 29.09 & 68.94 & 53.50 & 60.20 \\
CAKE & 50\% & 49.16 & 43.37 & 40.48  & 27.95 & 68.97 & \textbf{53.75} & 60.45 \\
\textbf{HeteroCache} & 50\% & \textbf{49.42} & \textbf{43.38} & \textbf{40.54} & 28.59 & \textbf{69.48} & \textbf{53.75} & \textbf{60.77} \\
\midrule
\rowcolor{baselinegray} \textit{Qwen2.5-14B-Instruct} & 100\% & 50.55 & 43.01 & 52.54 & 24.99 & 71.68 & 54.35 & 56.70 \\
ShadowKV & 30\% & 46.93 & 35.77 & 50.80 & 17.18 & 70.65 & 54.00 & 53.20 \\
OmniKV & 30\% & 46.48 & 37.90 & 50.21 & 23.10 & 68.44 & 50.25 & 48.95 \\
CAKE & 30\% & 46.92 & 40.74 & 42.95 & 23.69 & 66.77 & 54.25 & 53.09 \\
\textbf{HeteroCache} & 30\% & \textbf{49.97} & \textbf{42.46} & \textbf{52.12} & \textbf{24.12} & \textbf{71.18} & \textbf{54.63} & \textbf{53.33} \\
\bottomrule 
\end{NiceTabular}
}
\caption{Inference performance on LongBench \citep{bai-etal-2024-longbench}. \textit{Italics} indicate that the model uses FullAttention baseline. \textbf{Bold} indicates the best performance among compression methods.}

\label{tab:longbench}
\end{table*}

\begin{table*}[t]
\centering

\setlength{\tabcolsep}{9pt}
\setlength{\aboverulesep}{0pt} 
\setlength{\belowrulesep}{0pt}
\definecolor{baselinegray}{gray}{0.9}

\resizebox{\textwidth}{!}{%

    \begin{NiceTabular}{l | c | c | c c | c c c}[colortbl-like]
    \toprule

    & & & \multicolumn{2}{c|}{\textbf{Difficulty}} & \multicolumn{3}{c}{\begin{tabular}{c} \textbf{Length} \end{tabular}} \\

    \textbf{Methods} & \textbf{Mem\%} & \textbf{Overall} & \textbf{Easy} & \textbf{Hard} & \textbf{Short} & \textbf{Medium} & \textbf{Long} \\
    \midrule

    \rowcolor{baselinegray} \textit{Llama-3.1-8B-Instruct} & 100\% & 31.2 & 32.8 & 30.2 & 36.1 & 27.9 & 29.6 \\
    Quest                          & 100\%   & 28.0 & 29.7 & 27.0 & 33.3 & \textbf{27.4} & 20.4 \\
    ShadowKV                       & 50\%   & 30.0 & 30.2 & \textbf{29.9} & \textbf{36.1} & 26.5 & 26.9 \\
    OmniKV                         & 50\%    & 30.4 & 31.8 & 29.6 & \textbf{36.1} & 26.0 & 29.6 \\
    CAKE                           & 50\%    & 29.4 & 31.8 & 28.0 & 33.9 & 25.6 & 29.6 \\
    \textbf{HeteroCache}            & 50\%   & \textbf{30.6} & \textbf{32.3} & 29.6 & 35.6 & 26.0 & \textbf{31.5} \\
    \midrule
    \rowcolor{baselinegray} \textit{Qwen2.5-14B-Instruct} & 100\% & 33.2 & 35.4 & 31.8 & 42.8 & 29.3 & 25.0 \\
    ShadowKV                              & 30\%  & 33.0 &  \textbf{36.5} & 30.9 & 42.8 &  29.3 & 24.1 \\
    OmniKV                                & 30\%  & 33.3 & 36.2 & 31.4 & 43.6 & 28.5 & 26.0 \\
    CAKE                                  & 30\%  & 31.6 & 33.3 & 30.5 & 37.2 & \textbf{30.2} & 25.0 \\
    \textbf{HeteroCache}                  & 30\%  &  \textbf{33.6} & 35.9 &  \textbf{32.2} &  \textbf{44.4} & 27.4 &  \textbf{27.8} \\
    
    \midrule
    \rowcolor{baselinegray} \textit{DeepSeek-R1-Distill-Llama-8B} & 100\% & 29.2 & 32.8 & 27.0 & 32.8 & 25.1 & 31.5 \\
    Quest & 100\% & 24.1 & 30.2 & 20.3 & 23.3 & 25.1 & 23.1 \\
    ShadowKV & 50\% & 25.0 & 28.4 & 21.6 & 23.7 & 26.1 & 25.2  \\
    OmniKV & 50\% & 26.6 & 26.6 & 26.7 & 28.9 & \textbf{27.4} & 21.3 \\
    CAKE & 50\% & 27.2 & 28.6 & 26.4 & \textbf{33.9} & 26.5 & 17.6 \\
    \textbf{HeteroCache} & 50\% & \textbf{28.9} & \textbf{32.5} & \textbf{26.8} & 32.5 & 24.9 & \textbf{30.8} \\
    
    \bottomrule
    \end{NiceTabular}
}
\caption{Inference performance on LongBench v2 \citep{bai-etal-2025-longbench}. \textit{Italics} indicate that the model uses FullAttention baseline. \textbf{Bold} indicates the best performance among compression methods.}
\label{tab:longbenchv2}
\end{table*}

\section{Experiments}
\label{sec:experiments}
\subsection{Experiment Settings}

\paragraph{Backbone LLMs and Baselines.}
We selected multiple open-source LLMs for our experiments to verify the performance across different paradigms. These include standard non-reasoning models, specifically Llama-3.1-8B-Instruct and Qwen2.5-14B-Instruct, as well as the reasoning model DeepSeek-R1-Distill-Llama-8B.

To demonstrate the effectiveness of HeteroCache, we compare it with several compression methods:
(1) FullAttention. The original model without KV cache compression.
(2) Quest \citep{tang2024quest}. An algorithm that estimates the criticality of KV cache pages using the statistics of key values, loading only critical pages for attention calculation;
(3) ShadowKV \citep{sun2025shadowkvkvcacheshadows}. A dynamic compression method that retains low-rank key projections on the GPU while offloading the value cache to the CPU, dynamically reconstructing minimal sparse KV pairs on the fly;
(4) OmniKV \citep{hao2025omnikv}. A dynamic compression method by offloading KV pairs to CPU memory and retrieving relevant tokens based on layer attention patterns;
(5) CAKE \citep{qin2025cake}. A cascading static compression framework that allocates adaptive cache budgets between layers based on their spatiotemporal attention preferences.

\begin{table*}[t]
\centering

\setlength{\tabcolsep}{3.5pt}
\renewcommand{\arraystretch}{1.1} 

\definecolor{baselinegray}{gray}{0.9}

\resizebox{\textwidth}{!}{%
    \begin{NiceTabular}{l | c | c | c c c c c c c c c}[colortbl-like]
    \toprule
    \textbf{Methods} & \textbf{Mem\%} & \textbf{Avg.} & \textbf{En.Sum} & \textbf{En.QA} & \textbf{En.MC} & \textbf{En.Dia} & \textbf{Zh.QA} & \textbf{Code.Debug} & \textbf{Math.Find} & \textbf{Retr.PassKey} & \textbf{Retr.Number} \\
    \midrule
    
    \rowcolor{baselinegray} \textit{Llama-3.1-8B-Instruct} & 100\% & 47.56 & 32.93 & 27.41 & 65.94 & 19.50 & 34.60 & 25.38 & 24.00 & 99.66 & 98.64  \\
    Quest & 100\% & 46.68 & 28.86 & 26.55 & 64.00 & 18.50 & 31.78 & 24.65 & 22.86 & 99.15 & 98.01  \\
    ShadowKV & 50\% & 44.26 & 23.59 & 18.62 & 64.19 & 15.50 & 30.55 & \textbf{25.38} & 19.71 & 98.31 & 98.14  \\
    OmniKV & 50\% & 46.92 & 31.46 & \textbf{27.03} & \textbf{65.94} & 19.00 & 33.98 & 23.10 & \textbf{24.00} & \textbf{99.66} & 98.14  \\
    CAKE & 50\%  & 46.91 & 31.51 & 26.15 & \textbf{65.94} & 19.00 & 33.22 & 25.13 & \textbf{24.00} & \textbf{99.66} & 97.46 \\
    \textbf{HeteroCache} & 50\% & \textbf{47.34} & \textbf{32.05} & 26.94 & \textbf{65.94} & \textbf{19.50} & \textbf{34.14} & \textbf{25.38} & \textbf{24.00} & \textbf{99.66} & \textbf{98.47}   \\

    \bottomrule
    \end{NiceTabular}
}
\caption{Inference performance on InfiniteBench \citep{zhang-etal-2024-bench}.
\textit{Italics} indicate that the model uses FullAttention baseline. \textbf{Bold} indicates the best performance among compression methods.}
\label{tab:infinitebench}
\end{table*}

\paragraph{Implementation.} 
Detailed system configurations are provided in Appendix~\ref{sec:system_specs}. To demonstrate the orthogonality of HeteroCache with quantization, we applied 4-bit weight quantization to the DeepSeek-R1-Distill-Llama-8B model using bitsandbytes \citep{dettmers20218}.
To ensure a fair comparison, we standardized the evaluation criteria by requiring all algorithms to retain a prefetched KV cache equivalent to a memory budget ratio $\rho$, rather than a fixed token budget, and we used the Qwen2.5 series to ensure a fair comparison, as implementations of certain baselines do not yet support the latest architecture Qwen3. See Appendix~\ref{sec:system_memory} for detailed system footprint and bandwidth scaling.

\paragraph{Hyperparameter Configuration.} For the hyperparameter configuration, we standardized both the stability threshold $\tau_{\text{stable}}$ and the similarity threshold $\tau_{\text{sim}}$ at 0.5 for Llama and the stability threshold 0.4 for Qwen due to the lower memory budget. Consistent with this setting, the drift threshold $\tau_{\text{drift}}$ was set equal to $\tau_{\text{stable}}$. Regarding the KV cache memory budget $\rho$, we selected 0.3 or 0.5, tailored to the specific characteristics of different datasets and models. 
Ablation studies on parameters are presented in the following sections.
\begin{figure}[t!]
    \centering
    \includegraphics[width=\linewidth]{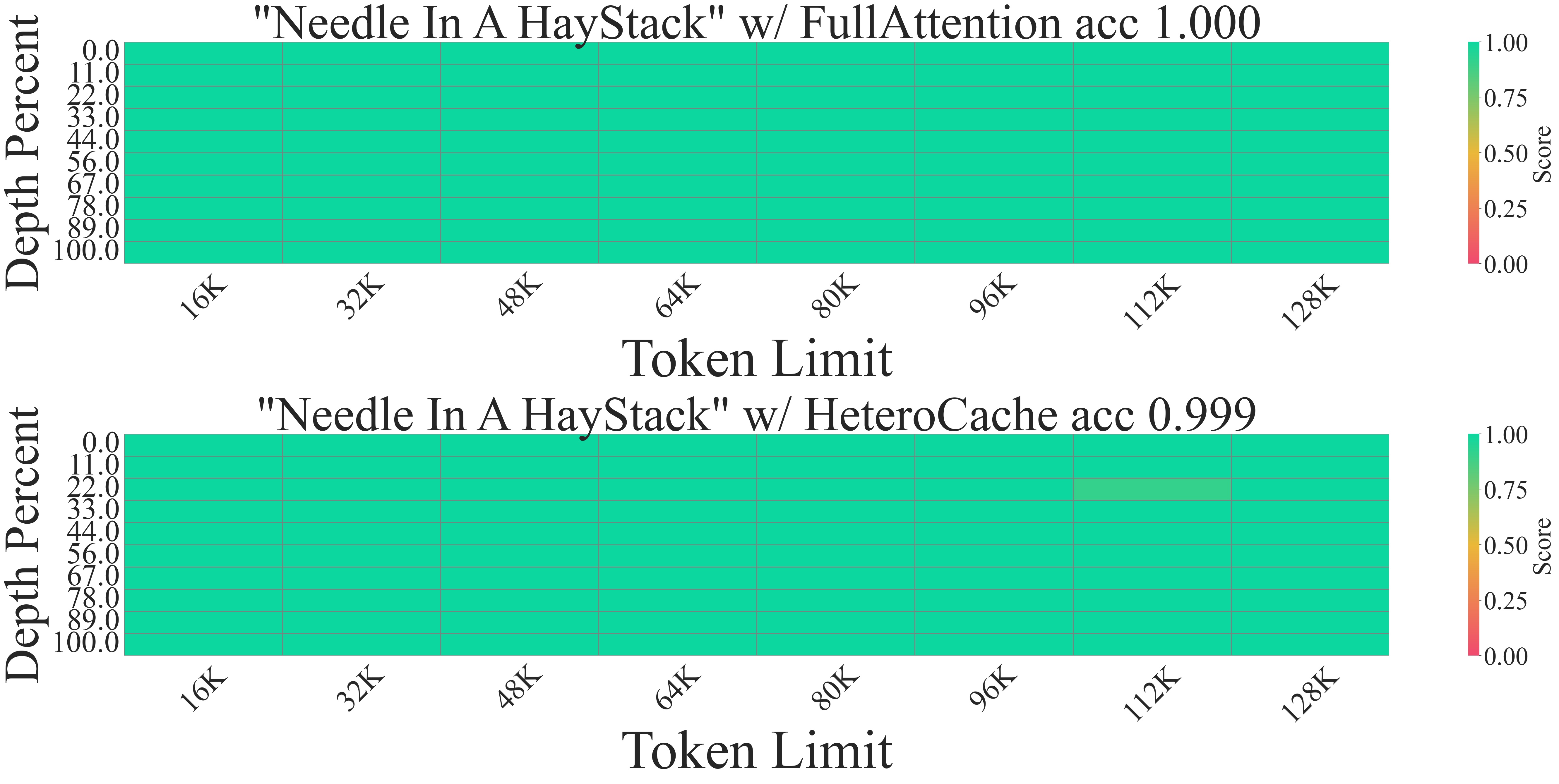}
    \setlength{\abovecaptionskip}{-0.3cm}
    \caption{Results for Llama-3.1-8B-Instruct on NIAH, evaluated on context lengths from 16K to 128K tokens.}
    \label{fig:needle}
\end{figure}

\subsection{Performance on Various Tasks}
\paragraph{Standard Benchmarks and Retrieval.}

To comprehensively validate the effectiveness of HeteroCache in long-context inference tasks, we conducted extensive experiments on LongBench and LongBench v2 \citep{bai-etal-2024-longbench,bai-etal-2025-longbench}. As summarized in Table~\ref{tab:longbench} and Table~\ref{tab:longbenchv2}, the results demonstrate that even under significantly compressed KV cache budgets, HeteroCache exhibits superior performance that is nearly indistinguishable from the FullAttention mechanism and significantly outperforms existing state-of-the-art baselines. This advantage is further substantiated by extending it to the more challenging InfiniteBench \citep{zhang-etal-2024-bench}. As shown in Table~\ref{tab:infinitebench}, HeteroCache maintains exceptional inference accuracy even with low memory budgets, achieving leading or highly competitive results across various subtasks, which fully proves its robustness in handling ultra-long context tasks.

This remarkable performance superiority stems from effectively overcoming the limitations of existing methods. Specifically, although ShadowKV and OmniKV also employ dynamic compression and offloading strategies, they primarily rely on coarse-grained compression for retrieval, which fails to capture fine-grained semantic dependencies and results in wasted cache space. In contrast, HeteroCache achieves more efficient information retention with a minimal memory footprint through fine-grained compression and retrieval at the head level. Regarding CAKE, although it rationally distributes cache resources based on layer importance, its lack of dynamic compression and recall mechanisms makes it unable to address attention drift during long-text generation. 
Furthermore, while Quest achieves certain acceleration effects through sparsity, results indicate that it fails to effectively reduce memory usage and instead leads to a noticeable drop in accuracy.
\begin{figure}[t!]
    \centering
    \includegraphics[width=\linewidth]{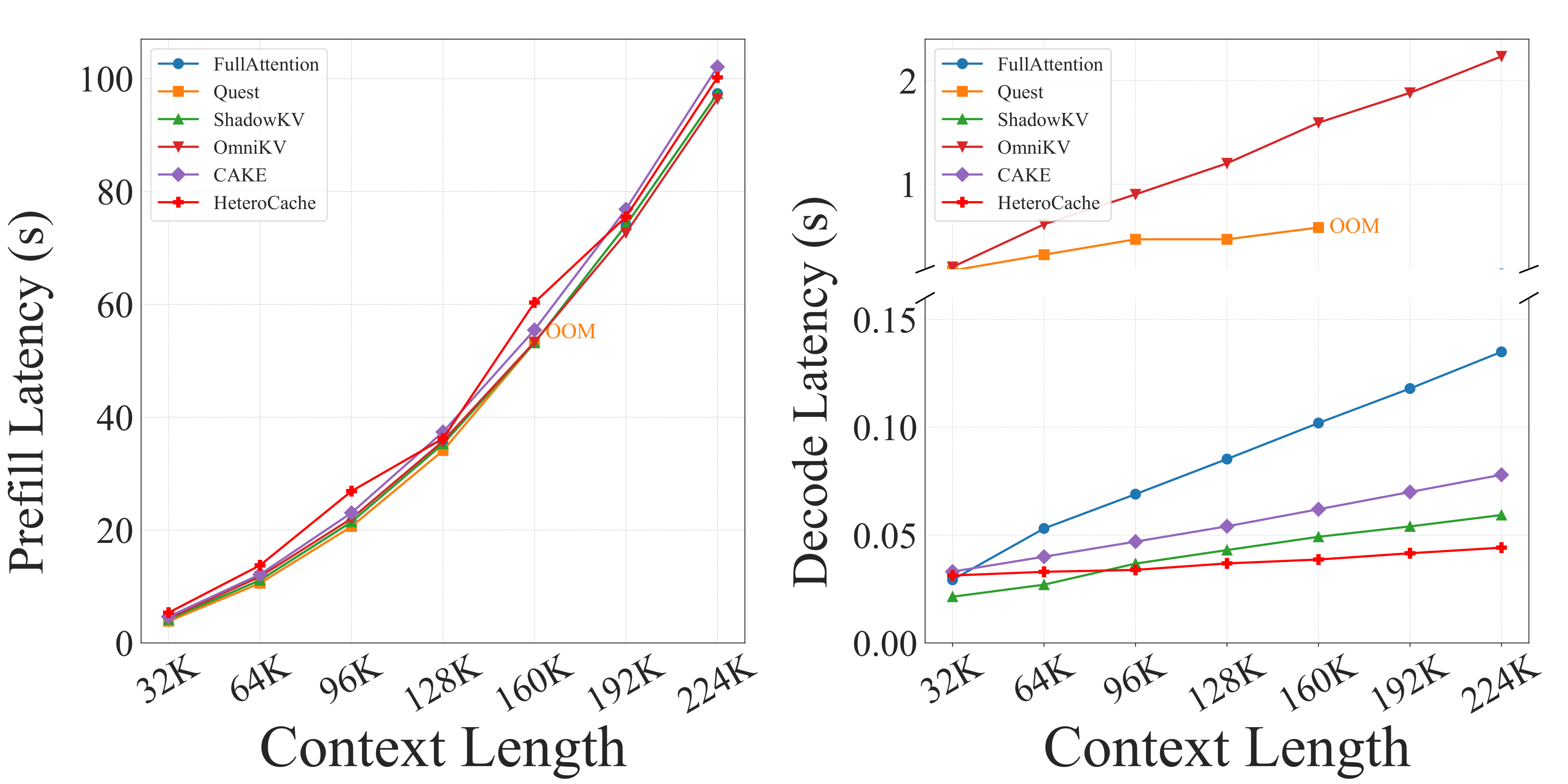}
    \setlength{\abovecaptionskip}{-0.3cm}
    \caption{End-to-end latency results for HeteroCache and the baselines. The left and right plots show the latency for the prefill and decode stage, respectively.}
    \label{fig:latency}
\end{figure}

Our performance on the Needle In A Haystack (NIAH) \citep{fu2024dataengineeringscalinglanguage} further corroborates this analysis. As illustrated in Figure~\ref{fig:needle}, HeteroCache achieves perfect retrieval accuracy across all context length settings from 16K to 128K, exhibiting robust information retention capabilities comparable to the FullAttention mechanism. Such consistent performance strongly demonstrates that our approach, through its fine-grained dynamic retrieval mechanism, can high-fidelity restore critical historical dependencies while significantly reducing GPU memory overhead, thereby providing reliable support for long-document question answering and complex reasoning tasks. Beyond standard single-needle retrieval, we evaluate the robustness of our method on U-NIAH \citep{10.1145/3786609}, detailed in the Appendix~\ref{sec:multi_needle}.

\paragraph{Reasoning Capabilities on DeepSeek-R1-Distill.} 
Emerging reasoning models \citep{Guo_2025} have improved their ability to solve complex problems through the chain of thought mechanism. However, the extremely long intermediate processes they generate pose a severe challenge to the efficiency of KV caching, making them an ideal scenario for testing dynamic retrieval capabilities. In the generation of long sequences, the loss of any key intermediate step may lead to the breaking of reasoning logic. As shown in Table~\ref{tab:longbenchv2}, our test results using the DeepSeek-R1-Distill-Llama-8B model on the LongBench v2 benchmark indicate that HeteroCache demonstrates significantly better performance than baselines under limited memory and is closest to the FullAttention mechanism. This confirms that its fine-grained dynamic strategy can precisely retain key thinking paths in long-distance reasoning, effectively ensuring the coherence of complex reasoning tasks under the condition of limited memory.

\begin{table}[t]
\centering

\resizebox{\linewidth}{!}{%
    \begin{tabular}{l cccc}
    \toprule
    \textbf{Methods} & \textbf{Mem\%} & \textbf{Avg.} & \textbf{Prefill (s)} & \textbf{Decode (s)} \\
    \midrule
    \textbf{HeteroCache} & 50\% & 46.00 & 2.30 & 0.032 \\
    
    w/o Allocation & 50\% & 44.97 & 2.30 & 0.035 \\
    
    w/o Retrieval & 50\% & 45.32 & 2.30 & 0.027 \\

    \bottomrule
    \end{tabular}%
}
\caption{Ablation study of the core components of HeteroCache on Qasper with Llama-3.1-8B-Instruct.}
\label{tab:ablation_component}  
\end{table}

\subsection{Latency}
We evaluated the end-to-end latency of HeteroCache against the baselines on the Llama-3.1-8B-Instruct model. For the prefill stage, we measured the Time-To-First-Token (TTFT), while for the decode stage, we calculated the average latency per token over 50 generated tokens.

The results presented in Figure \ref{fig:latency} illustrate the efficiency of the proposed method. In the prefill stage, although compression algorithms typically necessitate additional computations for metric statistics and cache selection, experimental results indicate that these overheads are minimal. Consequently, the HeteroCache TTFT curve overlaps with that of the FullAttention baseline, indicating that no significant latency bottleneck is introduced.
In the decode stage, HeteroCache shows exceptional performance. As the context length increases, it maintains consistently low inference latency, achieving approximately 3× acceleration compared to the FullAttention mechanism in the 224K context. More notably, in stark contrast to OmniKV which incurs substantial latency overhead due to frequent I/O operations, HeteroCache achieves a speedup of over 40× compared to OmniKV. Furthermore, starting from a context length of 128K, HeteroCache consistently outperforms all other compression algorithms, achieving the lowest per-token generation latency and verifying its superior efficiency and scalability in ultra-long context scenarios. 

Furthermore, to comprehensively evaluate the robustness of our asynchronous retrieval mechanism against potential I/O bottlenecks, we provide detailed analyses of the retrieval trigger rate, tail latency, and performance under constrained interconnects in the Appendix~\ref{sec:hardware_constraints}.

\begin{figure}[t!]
    \centering

    
    \includegraphics[width=\linewidth]{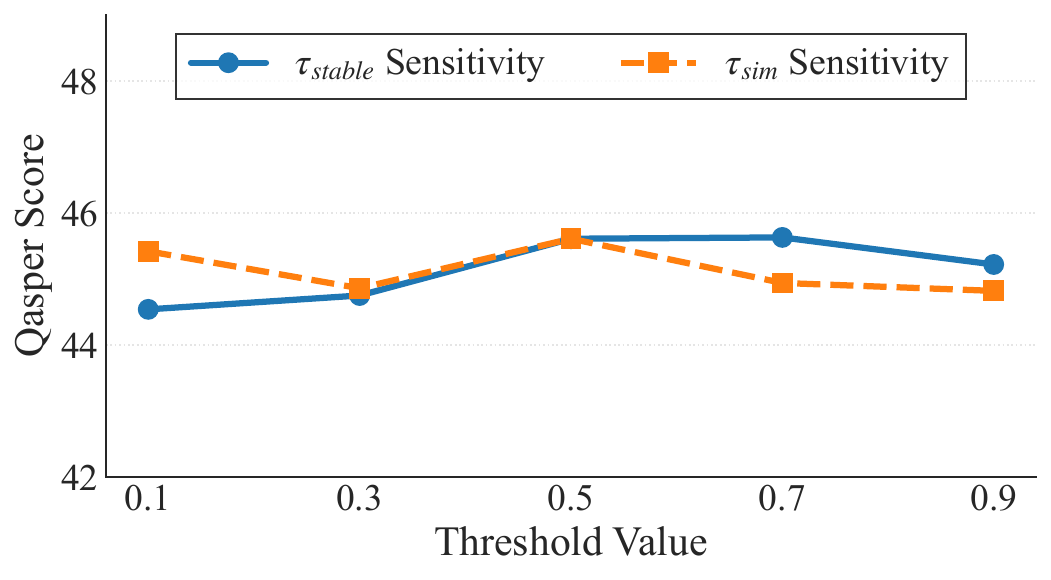}

    
    \caption{Ablation study of the threshold with Llama-3.1-8B-Instruct.} 
    \label{fig:sensitivity}

\end{figure}

\subsection{Ablation Studies}

To systematically validate the effectiveness of the core components within HeteroCache, we designed two key ablation variants and presented the results in Table \ref{tab:ablation_component}. We first examined the impact of the budget allocation strategy. The performance decline in the w/o Allocation variant, where a uniform cache budget is assigned, confirms that ignoring stability differences prevents compressed heads from retaining sufficient context. Subsequently, we evaluated the dynamic retrieval mechanism. The degradation in the w/o Retrieval setting verifies the existence of attention drift and highlights the necessity of pivot-guided updates for maintaining long-range reasoning. Furthermore, the integration of these components imposes a negligible latency overhead. Regarding the hyperparameters as illustrated in Figure \ref{fig:sensitivity}, our algorithm demonstrates robustness. Consequently, we selected the optimal value of 0.5 as the experimental threshold to balance the compression rate with the retention of information.

\section{Conclusion}
We propose HeteroCache, a training-free dynamic compression framework. By combining stability-based budget allocation with a head-guided asynchronous retrieval mechanism, HeteroCache effectively addresses the limitations of existing dynamic compression algorithms. It significantly outperforms prior methods in long-context tasks while drastically reducing the memory footprint. Furthermore, it achieves a $3\times$ decoding acceleration compared to the original model, offering an efficient and robust solution for resource-constrained inference.
\section*{Limitations}

Although HeteroCache demonstrates state-of-the-art performance in efficient long-context inference, we acknowledge a few limitations in the current implementation. First, our prototype is primarily based on high-level PyTorch operations to validate the algorithmic effectiveness. We have not yet implemented custom CUDA kernels for the sparse attention and dynamic retrieval modules. Consequently, the current speedup does not fully reflect the theoretical upper bound of our method, and further engineering optimizations could yield significant latency reductions. Second, as a heterogeneous storage framework, the efficiency of our asynchronous retrieval mechanism inherently relies on the interconnect bandwidth (e.g., PCIe) between the CPU and the GPU. While our supplementary experiments in Appendix~\ref{sec:hardware_constraints} demonstrate that HeteroCache maintains highly robust performance even on consumer-grade hardware with restricted bandwidth, extreme low-bandwidth environments might still constrain the ability to hide I/O latency completely. Future work will focus on developing optimized kernels and exploring hardware-aware prefetching strategies to fully mitigate these potential bottlenecks.
\section*{Ethical Considerations}

Our work introduces HeteroCache, a framework for compressing the KV cache in LLMs to reduce its memory overhead and accelerate the decoding process. The research is purely algorithmic in nature, with a primary focus on memory optimization. 

This study did not involve human subjects and all experiments were conducted on publicly available open-source models and standard academic benchmarks. No personal or private information was used. We do not foresee any direct ethical concerns arising from this work. We declare no conflict of interest.
\section*{Acknowledgments}
This work was supported in part by The National Nature Science Foundation of China (Grant Nos.: 62303406, 62432014), in part by Ningbo Key R\&D Program (No.: 2025Z055).

\bibliography{custom}

\appendix
\section*{Appendix}
\section{Pseudocode}
\label{sec:pseudocode}
To provide a clear perspective on our proposed method, we present the detailed HeteroCache workflow in the form of pseudocode. The entire step-by-step procedure is systematically illustrated in Algorithm \ref{alg:heterocache}.

\begin{algorithm*}[t!]
\caption{HeteroCache Inference Process}
\small
\label{alg:heterocache}
\begin{algorithmic}[1]
\REQUIRE Prompt $P$, Model $\mathcal{M}$, Thresholds $\tau_{\text{stable}}, \tau_{\text{sim}}, \tau_{\text{drift}}$, Budget Ratio $\rho$, Window $W$
\ENSURE Generated sequence $Y$

\STATE {// Stage 1: Head Profiling \& Taxonomy}
\FOR{each head $h \in \mathcal{H}$}
    \STATE Compute $S_{\text{stable}}^{(h)}$ and $S_{\text{sim}}^{(h)}$ via calibration
\ENDFOR
\STATE $\mathcal{H}_{\text{unique}} \leftarrow \{h \in \mathcal{H} \mid S_{\text{sim}}^{(h)} < \tau_{\text{sim}}\}; \quad \mathcal{H}_{\text{similar}} \leftarrow \{h \in \mathcal{H} \mid S_{\text{sim}}^{(h)} \ge \tau_{\text{sim}}\}$
\STATE $\mathcal{H}_{\text{anchor}} \leftarrow \{h \in \mathcal{H}_{\text{unique}} \mid S_{\text{stable}}^{(h)} \ge \tau_{\text{stable}}\}; \quad \mathcal{H}_{\text{volatile}} \leftarrow \{h \in \mathcal{H}_{\text{unique}} \mid S_{\text{stable}}^{(h)} < \tau_{\text{stable}}\}$
\STATE $\mathcal{C}_{\text{clusters}}=\{C_1, ..., C_m\} \leftarrow \text{GreedyStarClustering}(\mathcal{H}_{\text{similar}})$
\STATE $\mathcal{H}_{\text{pivot}} \leftarrow \{h_{\text{pivot}}^{(j)} \mid \forall C_j \in \mathcal{C}_{\text{clusters}}\}; \quad \mathcal{H}_{\text{satellite}} \leftarrow \bigcup_{j}(C_{j}\setminus\{h_{\text{pivot}}^{(j)}\})$

\STATE {// Stage 2: Stability-Based Budget Allocation}
\STATE $\mathcal{H}_{\text{comp}} \leftarrow \mathcal{H}_{\text{anchor}} \cup \mathcal{H}_{\text{satellite}}$
\STATE Calculate $L_{\text{base}} \leftarrow \frac{(\rho N - N_{\text{full}})L}{N_{\text{comp}}}$
\FOR{each head $h_i \in \mathcal{H}_{\text{comp}}$}
    \STATE $w_i \leftarrow 1/S_{\text{stable}}^{(i)}$
    \STATE $l_i \leftarrow L_{\text{base}} \cdot \frac{w_i}{\sum_{h_j \in \mathcal{H}_{\text{comp}}} w_j}$
\ENDFOR

\STATE {// Stage 3: Hierarchical Storage \& Dynamic Inference}
\STATE $\mathcal{C}_{\text{GPU}} \leftarrow \{\text{FullKV}^{(h)} \mid h \in \mathcal{H}_{\text{volatile}} \cup \mathcal{H}_{\text{pivot}}\} \cup \{\text{Top}_{l_i}(\text{KV}^{(h)}) \mid h \in \mathcal{H}_{\text{comp}}\}$
\STATE $\mathcal{C}_{\text{CPU}} \leftarrow \{\text{FullKV}^{(h)} \mid h \in \mathcal{H}_{\text{satellite}}\}$
\STATE $\mathcal{K}_{\text{base}}^{(p)} \leftarrow \text{TopIndices}_{L_{\text{base}}}(\text{Attn}_{\text{prefill}}^{(p)}) \quad \forall p \in \mathcal{H}_{\text{pivot}}$
\STATE $\mathcal{Q}_{\text{buffer}} \leftarrow \emptyset; \quad x_1 \leftarrow \text{LastToken}(P)$

\FOR{$t = 1$ \textbf{to} $T$}
    \STATE $y_t, q_t, \mathcal{C}_{\text{GPU}} \leftarrow \text{ModelForward}(\mathcal{M}, x_t, \mathcal{C}_{\text{GPU}})$
    \STATE $Y \leftarrow Y \cup \{y_t\}$
    \STATE $\mathcal{Q}_{\text{buffer}} \leftarrow \mathcal{Q}_{\text{buffer}} \cup \{q_t^{(p)} \mid p \in \mathcal{H}_{\text{pivot}}\}$ 
    
    \IF{$t \pmod W = 0$} 
        \FOR{each pivot head $h_p \in \mathcal{H}_{\text{pivot}}$}
            \STATE $\mathcal{W}_p \leftarrow \emptyset$
   
            \FOR{$k = t-W+1$ \textbf{to} $t$}
                \STATE $q_k^{(p)} \leftarrow \mathcal{Q}_{\text{buffer}}[k]$
                \STATE $\mathcal{K}_k^{(p)} \leftarrow \text{TopIndices}_{L_{\text{base}}}(\text{Attention}(q_k^{(p)}, \mathcal{C}_{\text{GPU}}))$
                \STATE $o_k \leftarrow |\mathcal{K}_k^{(p)} \cap \mathcal{K}_{\text{base}}^{(p)}| / L_{\text{base}}$
                \STATE $\mathcal{W}_p \leftarrow \mathcal{W}_p \cup \{o_k\}$
            \ENDFOR
            
            \STATE $r_t \leftarrow \mathbb{I}(\operatorname{Median}(\mathcal{W}_p) < \tau_{\text{drift}})$

            \IF{$r_t = 1$}
                \FOR{each $s \in \text{Cluster}(h_p) \cap \mathcal{H}_{\text{satellite}}$}
                    \STATE $\mathcal{I}_{s} \leftarrow \text{TopIndices}_{l_s}(\text{Attention}(q_t^{(p)}, \mathcal{C}_{\text{GPU}}))$
                    \STATE \textbf{AsyncUpdate}: $\mathcal{C}_{\text{GPU}}^{(s)} \leftarrow \mathcal{C}_{\text{CPU}}^{(s)}[\mathcal{I}_{s}]$
                \ENDFOR
                \STATE $\mathcal{K}_{\text{base}}^{(p)} \leftarrow \mathcal{K}_{t}^{(p)}$
            \ENDIF
        \ENDFOR
        \STATE $\mathcal{Q}_{\text{buffer}} \leftarrow \emptyset$
    \ENDIF
    \STATE $x_{t+1} \leftarrow y_t$
\ENDFOR
\RETURN $Y$
\end{algorithmic}
\end{algorithm*}

\section{Details of Observation}
\label{sec:appendix_b}

In this section, we provide a comprehensive description of the observation. This includes the data acquisition process and the methodology used to generate the motivational observations presented in Figure~\ref{fig:observation}.

\subsection{Data Collection and Preprocessing}
To construct a representative calibration dataset that ensures the generalization of our attention head taxonomy, we performed extensive web crawling from Wikipedia. We curated a diverse set of 50 samples that cover a wide range of domains, including science, entertainment, literature, and technology, to prevent topic-specific biases in attention patterns. Each collected sample was preprocessed and truncated to a standardized length of 10K tokens. This sequence length was chosen to be sufficiently long to trigger potential attention drift phenomena while remaining computationally manageable for the profiling phase. All subsequent metric calculations and profilings are based on the statistical average across this dataset to ensure robustness.
\begin{figure*}[t!]
    \centering
    \begin{subfigure}{0.48\textwidth}
        \centering
        \includegraphics[width=\linewidth]{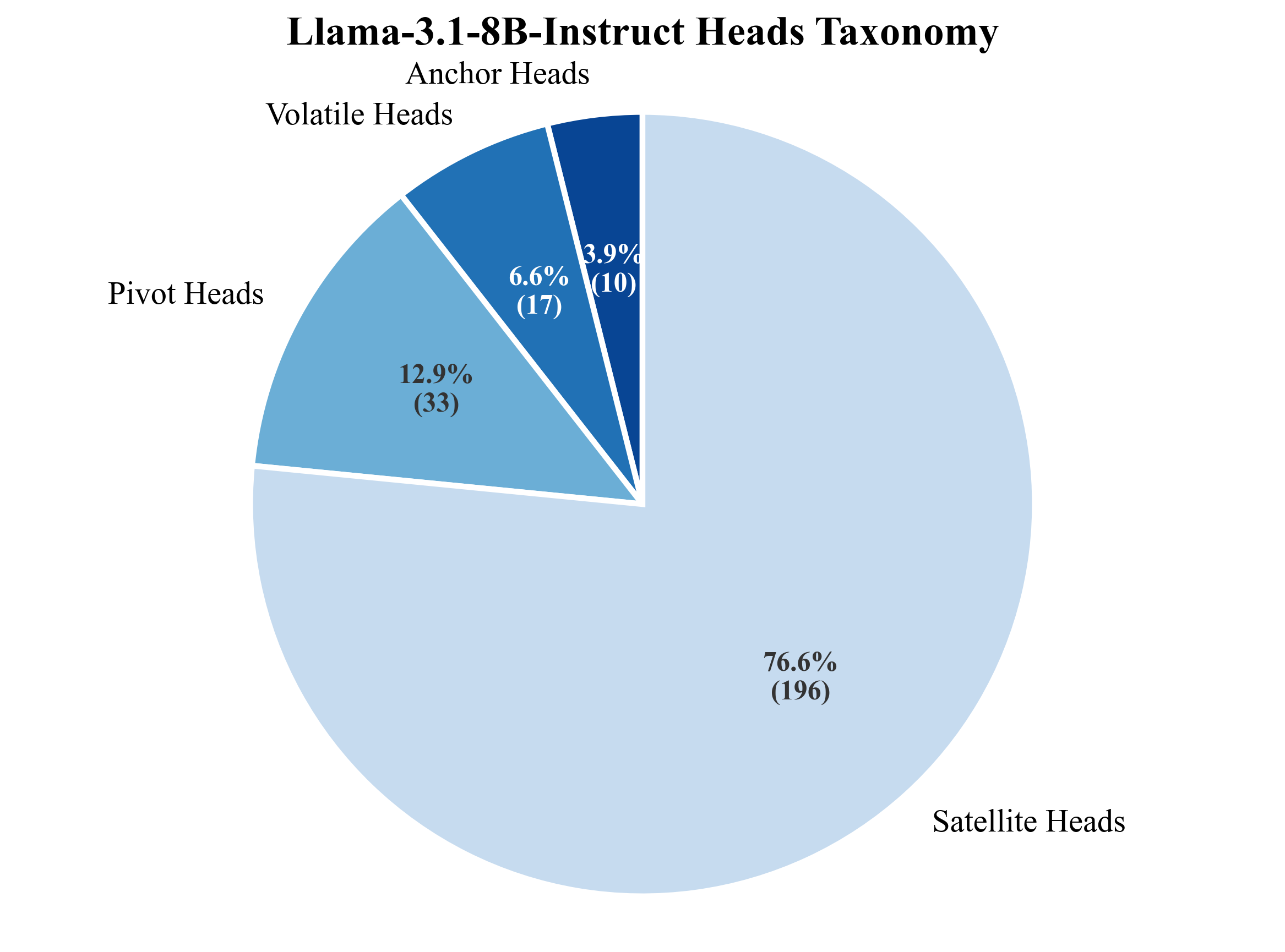} 
        \caption{Llama-3.1-8B-Instruct}
        \label{fig:llama3_pie}
    \end{subfigure}
    \hfill
    \begin{subfigure}{0.48\textwidth}
        \centering
        \includegraphics[width=\linewidth]{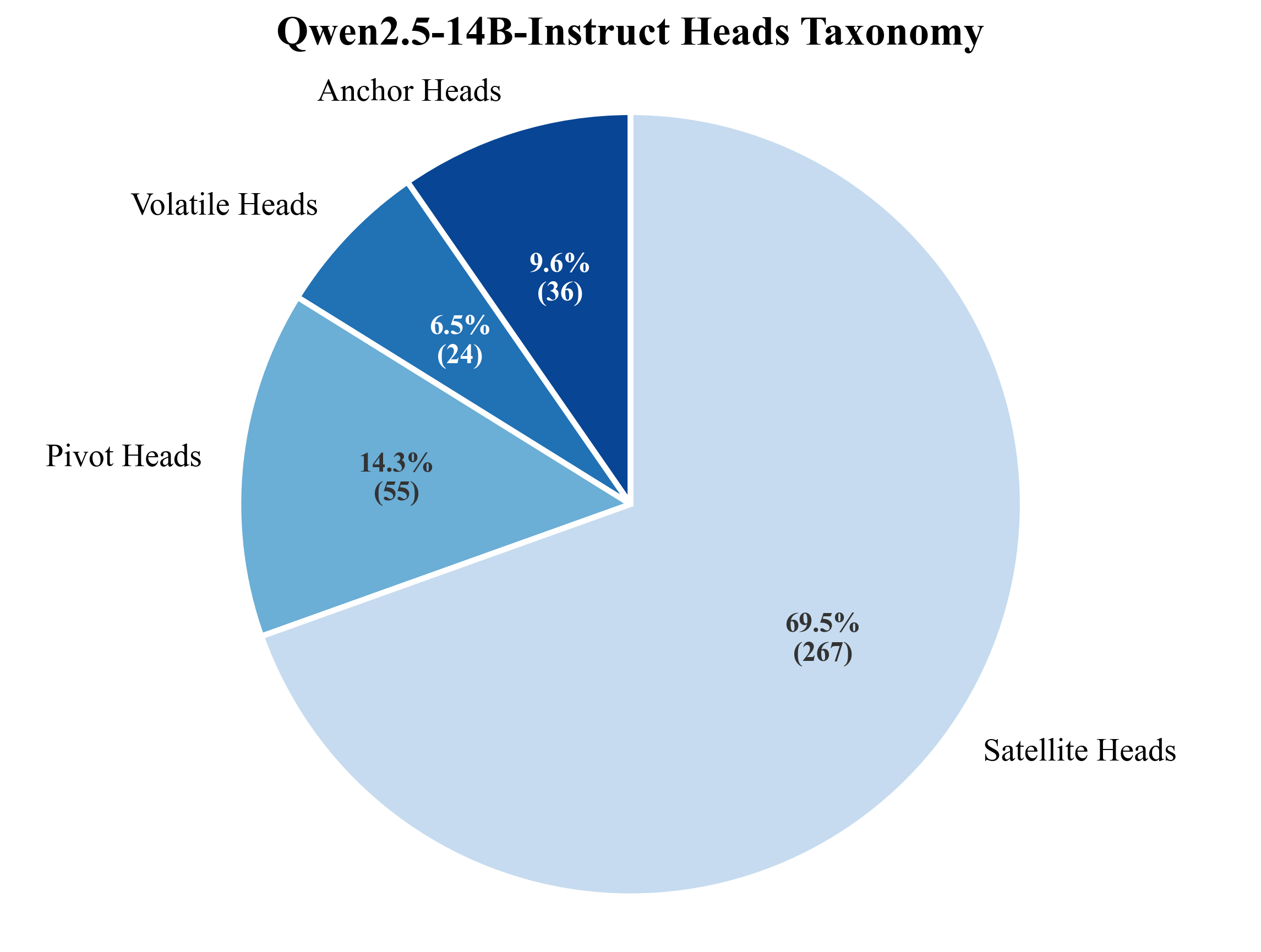} 
        \caption{Qwen2.5-14B-Instruct}
        \label{fig:qwen25_pie}
    \end{subfigure}
    \caption{The distribution of functional roles derived from our profiling algorithm. The dominance of anchor and satellite heads across both models demonstrates the generalizability of the HeteroCache taxonomy.}
    \label{fig:cluster_pie_charts}
\end{figure*}
\subsection{Analysis of Temporal Heterogeneity}
The temporal heterogeneity observed in Figure~\ref{fig:observation} (a) was generated using a specific experimental setup derived from our calibration dataset. We used the Llama-3.1-8B-Instruct model and selected a sample context of 10K tokens as input. The model then generated 100 subsequent tokens in a token-by-token decoding manner. To visualize the attention stability, we tracked two distinct heads: Layer 1 Head 5 as the representative stable head and Layer 2 Head 1 as the representative shifting head. For the calculation of the representative attention set, we implemented a rigorous extraction pipeline in each decoding step. Specifically, we first obtained the raw attention distribution of the last query token attending to all historical keys. To ensure the robustness of the selected indices against transient local noise and to capture the broader attention peaks, we applied average pooling operation motivated by the findings of
SnapKV \citep{li2024snapkv}. Following this smoothing process, we identified the indices corresponding to the top-1K attention scores to form the active token set $\mathcal{T}_k$. The overlap coefficient was then calculated at each step by measuring the intersection size between the $\mathcal{T}_k$ of the current decoding step and the $\mathcal{T}_k$ obtained during the prefill stage. A high overlap indicates that the head maintains its focus on the same historical context, whereas a declining overlap signifies attention drift.

\subsection{Analysis of Intralayer Redundancy}
To quantify the spatial redundancy visualized in the heatmap of Figure~\ref{fig:observation} (b), we extended the overlap analysis from individual heads to layer-level interactions. Utilizing the same smoothed Top-k extraction methodology described previously, we define the layer-level similarity as the average of the overlap coefficients from each head in a source layer to a target layer. This metric effectively captures the degree to which information in one layer is statistically correlated with that of another. The resulting visualization reveals that the diagonal blocks exhibit significantly higher values, confirming that heads within the same layer share a high degree of functional redundancy compared to cross-layer relationships.

\subsection{Head Classification and Thresholding}
The taxonomy of attention heads presented in Figure~\ref{fig:observation} (c) and Figure~\ref{fig:observation} (d) was derived by statistical analysis of the distribution of stability and similarity scores. For the stability classification shown in Figure~\ref{fig:observation} (c), we calculated the stability score $S_{\text{stable}}$ for each head by averaging its overlap coefficients throughout the recorded decoding trajectory. We then ranked all heads in descending order based on these scores. Applying a predefined stability threshold $\tau_{\text{stable}}$ of 0.5, we divide the distribution curve into two distinct regions. Heads with scores above this threshold were rendered in solid dark blue to designate stable heads, while those falling below were depicted with a dashed light blue line to represent shifting heads. Parallel to this, Figure~\ref{fig:observation} (d) illustrates the spatial similarity distribution. We computed the similarity score $S_{\text{sim}}$ for each head by measuring the maximum overlap it shares with any other head within the same layer at a representative decoding step. Similarly to the stability analysis, we ranked the heads according to their $S_{\text{sim}}$ values and applied a similarity threshold $\tau_{\text{sim}}$ of 0.5. The curve clearly differentiates between high similarity heads, which are redundant and suitable for clustering, and unique heads, which possess distinct attention patterns.

\section{Algorithmic Implementation Details}
\label{sec:appendix_c}

In this section, we provide supplementary algorithmic details for HeteroCache. We divide the process into preprocessing and metric computation, taxonomy and budget allocation, robustness analysis, and calibration transferability to illustrate the specific implementation logic.

\subsection{Preprocessing and Metric Computation}
The pipeline begins with the extraction and preprocessing of raw attention weights. For models employing grouped query attention, such as Llama-3.1 and Qwen2.5, we first aggregate the query attention scores by averaging the weights within the same KV group to align with the actual KV cache structure. To mitigate the impact of transient noise on the raw attention distribution, we apply a 1D average pooling operation to the attention scores of the last token before identifying significant entries. Based on these smoothed scores, we extract the top-$k$ indices to form the active token sets, which serve as the basis for calculating the stability and similarity metrics. Specifically, the stability score is computed as the median of the historical overlap coefficients between the decoding steps and the prefill stage to ensure robustness against outliers. Simultaneously, we construct a layer-wise adjacency graph by establishing edges between heads whose pairwise overlap ratio exceeds the predefined similarity threshold.

\subsection{Taxonomy and Budget Allocation}
Building on the constructed adjacency graph and stability metrics, we execute the taxonomy using a greedy star clustering algorithm. The process iterates through all attention heads to identify their functional roles. We first calculate the effective degree for each unassigned head, which represents the count of its neighbors that are also currently unassigned. In each iteration, the head with the highest effective degree is selected as a pivot head, and all its unassigned neighbors are immediately grouped in its cluster as satellite heads. This greedy strategy effectively maximizes the coverage of redundant structures. Upon completion of clustering, any remaining heads that were not assigned to a cluster are classified based on their temporal behavior. Those with stability scores exceeding the threshold are labeled as anchor heads, while the few remaining unstable heads are designated as volatile heads. Following this classification, we determine the cache budget for the compressible subset which includes both anchor and satellite heads. We assign a weight to each head that is inversely proportional to its stability score. These weights are then normalized to derive the final cache allocation ratio to ensure that heads prone to context shifts receive a larger share of the budget.

\subsection{Robustness Analysis}
To empirically verify the robustness and generalizability of our proposed taxonomy, we applied this profiling pipeline to two distinct mainstream model architectures, including Llama-3.1-8B-Instruct and Qwen2.5-14B-Instruct. Regarding the hyperparameter configuration, we standardized both the stability and similarity thresholds at 0.5 for the Llama-3.1 model. For Qwen2.5, we utilized a stability threshold of 0.4 and a similarity threshold of 0.5 to accommodate its specific attention dynamics. The resulting distributions of the four functional roles are visualized in Figure \ref{fig:cluster_pie_charts}. Consistent with our preliminary observations, the classification results for both model families reveal a striking commonality where the vast majority of attention heads are categorized as satellite heads. These heads exhibit high redundancy and are suitable for aggressive compression. Crucially, the proportion of volatile heads remains minimal across both architectures. This pervasive pattern confirms that the sparsity of attention and head redundancy are intrinsic characteristics shared by most mainstream LLMs, rather than being model-specific quirks. Consequently, our approach demonstrates strong robustness by guaranteeing that retaining the full KV cache for these highly dynamic heads does not occupy a substantial portion of the memory budget while maintaining high model performance.

\subsection{Calibration and Transferability}
\label{app:calibration_efficiency}

\paragraph{Calibration Cost.} The offline calibration process is highly efficient. It requires only a single forward pass along with minimal statistical computation. Empirically, the entire clustering process can be completed in a few minutes, introducing negligible overhead prior to deployment. Since the profiling pipeline only needs to extract and aggregate attention weights rather than storing massive intermediate hidden states, the entire process easily fits within a single consumer-grade GPU.

\paragraph{Sensitivity and Transferability.} To evaluate sensitivity to calibration data and transferability to longer contexts, we conducted additional experiments using the LongBench Pro \citep{chen2026longbenchprorealisticcomprehensive}, which consists of real natural long documents. We selected twenty samples each from three distinct tasks: Version and Code Diff Analysis, Structured and Numeric Reasoning, and Dialogue Memory and Long Horizon Tracking. 

To prove that offline clustering is robust to extended input lengths, we evaluated all these samples using a 128K context length. We used the Llama-3.1-8B-Instruct model and set the top-$k$ threshold to one-fifth of the length, keeping all other configurations identical to those in the paper. The clustering results of these three different tasks show that the newly identified satellite heads overlap with our original clustering results by more than 80\%. This demonstrates that our head taxonomy is highly robust, and the stability and redundancy patterns hold consistently at the 100K scale and beyond.

Furthermore, to ensure that this structural consistency does not significantly affect downstream accuracy, we tested the model's performance on the Qasper dataset from LongBench using these new task-specific clustering configurations. As shown in Table \ref{tab:qasper_transferability}, these consistent results confirm that our calibration is effectively transferred to extremely long contexts and diverse domains without performance degradation.

\begin{table}[h]
\centering
\resizebox{\columnwidth}{!}{%
\begin{tabular}{lc}
\toprule
\textbf{Task (Calibration Source)} & \textbf{Qasper Score} \\
\midrule
Version \& Code Diff Analysis & 45.72 \\
Structured \& Numeric Reasoning & 45.61 \\
Dialogue Memory \& Long-Horizon Tracking & 45.68 \\
\bottomrule
\end{tabular}%
}
\caption{Performance on the Qasper dataset using clustering configurations derived from different 128K LongBench Pro tasks.}
\label{tab:qasper_transferability}
\end{table}

\section{Additional Experiments and Analyses}
\label{sec:additional_experiments}

\subsection{Detailed System Specifications}
\label{sec:system_specs}

To ensure reproducibility and facilitate fair comparison across different hardware environments, we provide the complete system configuration used for our experimental evaluations in Table \ref{tab:system_specs}. Our analysis emphasizes the importance of the memory hierarchy and interconnect bandwidth, as these factors significantly influence the efficiency of the asynchronous KV cache retrieval mechanism.

\begin{table}[h]
\centering
\small
\resizebox{\columnwidth}{!}{%
\begin{tabular}{ll}
\toprule
\textbf{Component} & \textbf{Specification} \\
\midrule
CPU & Intel® Xeon® Gold 6338 @ 2.00 GHz \\
CPU RAM & 256 GiB DDR4 \\
GPU & NVIDIA A100 PCIe 80GB \\
GPU Memory & 80 GB HBM2e \\
GPU Memory Bandwidth & 1,935 GB/s \\
PCIe Interface & PCIe Gen4 x16 \\
Theoretical PCIe Bandwidth & $\approx$ 31.5 GB/s (per direction) \\
Interconnect & PCIe 4.0 \\
\bottomrule
\end{tabular}}
\caption{Detailed hardware and system configuration for experimental evaluation.}
\label{tab:system_specs}
\end{table}
\subsection{Memory and Bandwidth Profiling}
\label{sec:system_memory}

To provide a holistic view of the system's resource consumption, we measured the peak CPU memory usage and the corresponding PCIe bandwidth utilization during inference.

As summarized in Table \ref{tab:cpu_memory_scaling}, the peak CPU RAM usage scales linearly with the context length. For a 128K context, the CPU memory footprint reaches 17.95 GB, which remains well within the capacity of standard modern servers. Furthermore, peak bandwidth utilization reaches 21.07 GB/s, which is safely below the theoretical PCIe 4.0 limit. These results demonstrate that HeteroCache is highly scalable and does not introduce severe system-level memory or I/O bottlenecks when extending to long contexts.

\begin{table}[h]
\centering
\small
\resizebox{\columnwidth}{!}{%
\begin{tabular}{lcc}
\toprule
\textbf{Context Length} & \textbf{CPU Memory Peak} & \textbf{Bandwidth Utilization} \\
\midrule
32K  & 5.85 GB & 16.91 GB/s \\
64K  & 9.88 GB & 19.64 GB/s \\
128K & 17.95 GB & 21.07 GB/s \\
\bottomrule
\end{tabular}
}
\caption{Scaling of peak CPU memory footprint and PCIe 4.0 bandwidth utilization across different context lengths.}
\label{tab:cpu_memory_scaling}
\end{table}

\subsection{Multi-Needle Retrieval Evaluation}
\label{sec:multi_needle}

While the standard NIAH evaluation demonstrates HeteroCache's fundamental capability to retain critical information, it primarily assesses single-needle retrieval and may not fully reflect the complexities of multi-hop reasoning or the presence of adversarial distractors. To address this and avoid overgeneralizing single-needle performance, we conducted additional experiments using the more challenging U-NIAH \citep{10.1145/3786609} for multi-needle evaluation.

Following the exact same configuration for the Llama-3.1-8B-Instruct model used in our main experiments, we measured the performance using the ROUGE-1 metric across various context lengths. As shown in Table \ref{tab:uniah_results}, HeteroCache maintains a highly competitive performance compared to the FullAttention baseline. These results confirm that our dynamic retrieval mechanism successfully preserves complex, reasoning-centric dependencies in long-context environments.

\begin{table}[h]
\centering
\small
\begin{tabular}{lcccc}
\toprule
\textbf{Method} & \textbf{16K} & \textbf{32K} & \textbf{64K} & \textbf{128K} \\
\midrule
FullAttention & 7.379 & 3.338 & 1.306 & 2.195 \\
HeteroCache   & 7.312 & 3.327 & 1.213 & 2.195 \\
\bottomrule
\end{tabular}
\caption{ROUGE-1 scores on the multi-needle U-NIAH benchmark across different context lengths.}
\label{tab:uniah_results}
\end{table}

\subsection{Detailed Latency Analysis}
\label{sec:hardware_constraints}

\paragraph{Retrieval Frequency and Tail Latency.} 
A potential concern with dynamic retrieval mechanisms is the risk of excessive CPU-GPU communication causing severe latency spikes. However, our fine-grained tracking effectively minimizes unnecessary data fetching. To comprehensively evaluate performance during long context reasoning, we extended the testing window to a continuous generation of 1K tokens. In this configuration, the retrieval-trigger rate is only 0.823\%. 

To further verify the stability of the generation, we report the latency of the tail (P95 and P99) across various context lengths in Table \ref{tab:tail_latency}. The results indicate that the worst-case decoding steps remain highly efficient. These results thoroughly demonstrate that the system maintains an extremely low stall rate and stable low latency even during prolonged generation periods, ensuring that the asynchronous retrieval mechanism does not introduce severe stalling.

\begin{table}[h]
\centering
\small
\begin{tabular}{lcc}
\toprule
\textbf{Context Length} & \textbf{P95 Latency (s)} & \textbf{P99 Latency (s)} \\
\midrule
32K  & 0.068 & 0.074 \\
64K  & 0.092 & 0.108 \\
96K  & 0.095 & 0.114 \\
128K & 0.116 & 0.123 \\
160K & 0.118 & 0.126 \\
192K & 0.090 & 0.127 \\
224K & 0.096 & 0.132 \\
\bottomrule
\end{tabular}
\caption{Tail decoding latency (P95 and P99) across various context lengths.}
\label{tab:tail_latency}
\end{table}

\paragraph{Performance under Constrained Interconnects.} 
Whether the CPU$\leftrightarrow$GPU fetch overhead can be fully hidden depends heavily on the interconnect bandwidth. To validate the practicality of HeteroCache under hardware-constrained scenarios, we evaluated its performance on a consumer-grade setup featuring an NVIDIA GeForce RTX 4090 GPU and an Intel Xeon Gold 5218R processor. This setup utilizes a PCIe Gen 3 x16 interface, which restricts the theoretical bidirectional bandwidth to approximately 15.75 GB/s. 

As shown in Table \ref{tab:constrained_hardware}, despite the restricted interconnect bandwidth, HeteroCache maintains prefill and decode times that are nearly identical to the FullAttention baseline across various sequence lengths. This demonstrates that its asynchronous, infrequent retrieval strategy effectively mitigates I/O contention, allowing it to accelerate inference even on standard consumer-grade hardware.

\begin{table}[h]
\centering
\resizebox{\columnwidth}{!}{%
\begin{tabular}{lcccc}
\toprule
\textbf{Method} & \textbf{4K (Pre/Dec)} & \textbf{8K (Pre/Dec)} & \textbf{16K (Pre/Dec)} & \textbf{32K (Pre/Dec)} \\
\midrule
FullAttention & 0.497s / 0.038s & 1.169s / 0.041s & 2.496s / 0.039s & 4.943s / 0.042s \\
CAKE          & 0.624s / 0.041s & 1.332s / 0.038s & 2.757s / 0.037s & 5.008s / 0.039s \\
HeteroCache   & 0.502s / 0.039s & 1.185s / 0.040s & 2.508s / 0.038s & 4.951s / 0.039s \\
\bottomrule
\end{tabular}%
}
\caption{Prefill and Decode latency evaluated on NVIDIA GeForce RTX 4090.}
\label{tab:constrained_hardware}
\end{table}

\section{Statement on the Use of LLMs}
\label{sec:appendix_llm}

We report on the use of LLMs in the preparation of this paper. The use of LLMs was strictly limited to the role of a general-purpose writing assistant. Specifically, we used these tools to proofread, correct grammatical errors, and rephrase sentences to improve clarity and readability. The LLMs did not contribute to the core scientific aspects of this work, such as research ideation, experimental design, or the generation of substantive content.

\end{document}